\documentclass[preprint,12pt]{elsarticle}

\usepackage[T1]{fontenc}
\usepackage[utf8]{inputenc}
\usepackage{amsmath,amssymb}
\usepackage{graphicx}
\usepackage{booktabs}
\usepackage{multirow}
\usepackage{xcolor}
\usepackage[hidelinks]{hyperref}
\usepackage{subcaption}
\usepackage{siunitx}
\usepackage{placeins}

\newcommand{\mAP}{mAP@0.5:0.95}
\newcommand{\mAPf}{mAP@0.5}
\newcommand{\Emac}{$E_\text{MAC} = \SI{4.6}{pJ}$}
\newcommand{\Eac}{$E_\text{AC} = \SI{0.9}{pJ}$}

\begin{document}

\begin{frontmatter}

\title{Spiking Neural Networks for Energy-Efficient Object Detection
       in Forward-Looking Sonar Imagery}

\author[ece]{Gwenevere Frank}
\ead{jfrank@ucsd.edu}
\author[be]{Gert Cauwenberghs}

\address[ece]{Department of Electrical Engineering, University of California San Diego,
La Jolla, CA 92093, USA}
\address[be]{Department of Bioengineering, University of California San Diego,
La Jolla, CA 92093, USA}

\begin{abstract}
Autonomous underwater vehicles (AUVs) are increasingly important tools in industries
ranging from research, to energy, to defense. AUVs are power-constrained platforms operating
in remote environments with fixed battery capacities, where propulsion competes with compute
and sensors for power over lengthy mission durations. AUVs frequently operate in dark or
turbid waters where optical sensing is of limited value, and rely on sonar as their primary
sensing modality. Convolutional neural networks (CNNs) are the state-of-the-art solution
for object detection in forward-looking sonar imagery, but are energy expensive (e.g.\ YOLOv8m: 322~mJ/inference). Spiking neural networks (SNNs) rely on binary spike activations and
thus sparse accumulate-only operations, allowing them to be remarkably energy efficient,
particularly when paired with dedicated neuromorphic hardware. The sparse, high-contrast
structure of forward-looking sonar (FLS) returns is structurally matched to spike coding
in a way that optical imagery is not. No prior work has assessed the suitability of SNNs
for object detection in FLS imagery. SpikeYOLO~\cite{spikeyolo2024}, a fully spiking network trained with surrogate gradients, was benchmarked against state-of-the-art CNN baselines on three FLS object detection datasets. Key results:
\begin{itemize}
  \item SpikeYOLO T=2 achieves 3.3$\times$ lower theoretical compute energy on UATD (97 vs 322~mJ) at competitive accuracy (0.529~mAP@0.5:0.95 vs.\ YOLOv8m's 0.575).
  \item SpikeYOLO matches YOLOv8m on mAP@0.5 and outperforms YOLO-SONAR and Fast R-CNN baselines on the sparse Marine-Debris-FLS dataset at 4.4$\times$ lower energy.
  \item SpikeYOLO demonstrates superior robustness to multiplicative speckle noise (3.0\% degradation at $\sigma{=}0.4$ vs.\ 8.9\% for YOLOv8m), outperforming YOLOv8m outright at $\sigma{=}0.6$, directly relevant to real-world FLS deployment.
\end{itemize}
\end{abstract}


\end{frontmatter}

\section{Introduction}
\label{sec:intro}

Autonomous underwater vehicles (AUVs) are highly power-constrained platforms. AUVs are typically powered entirely from an internal battery that must power both the propulsion system and the compute and sensor suite (referred to as the hotel load) for often lengthy missions. Survey-class AUVs, such as the REMUS~600~\cite{remus600whoi} and Bluefin-21~\cite{bluefin21gdms}, carry 16-17~kWh of battery capacity. Hotel loads draw tens to hundreds of watts~\cite{williams2010endurance}, becoming a non-trivial fraction of battery capacity over long-duration missions.

AUVs executing survey, inspection, and search-and-rescue missions must sustain object detection throughout the full dive: obstacle avoidance during transit~\cite{petillot2001obstacle}, mine detection in mine-countermeasure (MCM) operations~\cite{palomeras2022mcm}, victim localization in maritime search-and-rescue~\cite{hu2024rescue}, and benthic habitat characterization~\cite{williams2012benthic}, all across dives routinely exceeding 10~hours~\cite{remus600whoi,bluefin21gdms,williams2010endurance}. State-of-the-art detectors such as RT-DETR-L~\cite{rtdetr} are energy-expensive, consuming hundreds of millijoules per inference.

In the turbid, unlit conditions typical of many AUV operations, optical sensing is of limited utility~\cite{uatd2022,karimanzira2020}. Forward-looking sonar (FLS) addresses this by providing real-time acoustic imagery of objects ahead of the vehicle, making it the standard sensor for the obstacle avoidance and detection tasks required of AUVs. FLS imagery is often sparse, high-contrast, and edge-dominated, lacking the texture and color of optical imagery. FLS presents its own challenges for object detection: speckle noise, acoustic shadowing, multipath reverberation, and a scarcity of labeled data~\cite{karimanzira2020,uatd2022}. CNNs such as YOLOv8m~\cite{yolov8} and transformers such as RT-DETR-L~\cite{rtdetr} are currently the state of the art for object detection in FLS imagery.

Spiking neural networks (SNNs) are often referred to as the third generation of neural
networks~\cite{maass1997}. Rather than propagating information as continuous floating-point
values, SNNs propagate data as sparse binary spike events. Because spike activations are
binary, synaptic inputs reduce to accumulate (AC) operations, adding a synaptic weight
to the membrane potential when a spike arrives, replacing the multiply-accumulate (MAC)
operations of traditional ANNs. AC operations cost roughly 5$\times$ less energy than MACs
in 45~nm CMOS~\cite{horowitz2014}, and the sparse firing of biologically-inspired neurons
amplifies this advantage further.

The sparse, high-contrast nature of FLS imagery naturally produces sparse spike patterns,
activating only a subset of neurons per timestep and amplifying the per-operation energy
advantage, suggesting a match between FLS sonar and SNN computation not
present in optical imagery. Spiking neural networks have been applied to object detection
in optical imagery, including ANN-to-SNN converted architectures such as
Spiking-YOLO~\cite{kimspikingyolo2019} and networks trained directly via surrogate gradient
descent such as EMS-YOLO~\cite{su2023emsyolo} and SpikeYOLO~\cite{spikeyolo2024}.
SU-YOLO~\cite{suyolo2025} extended this to underwater optical imagery, which retains the
texture and color cues absent in sonar. To date, however, no study has attempted to apply
SNNs to object detection in sonar imagery.

This paper makes the following contributions:
\begin{enumerate}
  \item The first evaluation of SNNs for object detection in FLS sonar imagery, benchmarked
        across three datasets (UATD~\cite{uatd2022}, Marine-Debris-FLS~\cite{valdenegro2025marine},
        WHFLS~\cite{wangwhfls2025}) against strong baselines including YOLOv8m~\cite{yolov8} and RT-DETR-L~\cite{rtdetr}.
  \item SpikeYOLO~\cite{spikeyolo2024} T=2 achieves 3.3$\times$ lower theoretical compute energy than YOLOv8m on UATD (97 vs.\ 322~mJ) at competitive accuracy (0.529 vs.\ 0.575~\mAP{}).
  \item Speckle-noise robustness evaluation: SpikeYOLO degrades 3.0\% vs.\ 8.9\% for
        YOLOv8m at $\sigma{=}0.4$, and outperforms YOLOv8m at $\sigma{=}0.6$, with LIF threshold gating providing structural immunity
        to multiplicative noise.
  \item A theoretical energy model~\cite{horowitz2014} quantifying inference
        energy across all architectures.
\end{enumerate}

\FloatBarrier
\section{Related Work}
\label{sec:related}

\FloatBarrier
\subsection{Object Detection in Forward-Looking Sonar}
Despite often being the preferred imaging modality for AUVs, sonar presents several properties that complicate object detection~\cite{karimanzira2020,dossantos2017}. FLS images have inhomogeneous resolution, with regions further from the vehicle imaged at lower resolution than nearby regions. Multipath acoustic returns can produce duplicate echoes of the same object, while objects blocking acoustic propagation create acoustic shadows (dark regions with no return) that can mimic or occlude targets. Objects equidistant from the sonar but at different heights collapse to the same location in the two-dimensional image. Acoustic wave interference produces spatially varying speckle noise throughout the image, and multiple scattering effects combine to produce non-uniform intensity across the scene.

Despite these challenges, convolutional neural networks have established a strong track record for object detection in FLS imagery. Karimanzira et al.~\cite{karimanzira2020} demonstrated successful detection of AUV docking station components using a Fast R-CNN architecture, training on water-tank images collected with two sonar devices and achieving average precision of 0.752--0.881, with inference deployed to an NVIDIA Jetson at 0.52~s per frame. Valdenegro-Toro~\cite{valdenegro2016} introduced the first end-to-end CNN for FLS object detection: a sliding-window architecture processing 96$\times$96 pixel crops, with shared convolutional features feeding separate detection and classification sub-networks trained jointly via a dual-objective loss. Evaluated on a nine-class water-tank dataset of 1300 images, the system achieved 93\% detection recall and 75\% classification accuracy.

More recent work has brought modern architectures to FLS. Wang et al.~\cite{wangwhfls2025} proposed a YOLOv7-based detector augmented with coordinate attention, spatial group-enhance attention modules, and a context feature extraction module designed for the sparse, high-contrast structure of FLS imagery. YOLOv8 has been adapted for FLS sonar with attention mechanisms and additional detection heads, achieving 98.2\% mAP@0.5 on the URPC2021 competition dataset~\cite{zheng2024yolov8sonar}. Transformer-based detectors have further pushed accuracy: RT-DETR~\cite{rtdetr} reaches 0.647~\mAP{} on UATD, and has been adapted for underwater sonar with enhanced attention and feature fusion modules (US-DETR~\cite{usdert2025}), achieving 95.3\% mAP@0.5 on the URPC2021 dataset (a 6000-image, 8-class subset drawn from the same UATD corpus), marginal gains in mAP@0.5:0.95 over the vanilla model (+0.3pp). None of these works, however, considers inference energy as a deployment constraint. All assume GPU-based inference rather than the power-limited embedded compute available on AUVs.

\FloatBarrier
\subsection{Energy-Constrained Perception on AUVs}
Power is one of the most precious resources an AUV has and it must be managed carefully. Cruz~\cite{cruz2019}, Bradley et al.~\cite{bradley2001auv}, and Williams~\cite{williams2010endurance} show in their formulations that the maximum range of an AUV and its maximum mission duration are both directly affected by its hotel load (the power consumed by the non-propulsion components of the AUV, including computers and sensors). It is quite typical for a survey AUV to have a hotel load of around 50~W while transiting to its location and around 100--200~W while it is performing a survey of the ocean floor.

FLS object detection has been deployed to embedded GPU platforms such as the NVIDIA Jetson Xavier NX~\cite{karimanzira2020}, which draws 10--15~W~\cite{jetsonnx} representing a non-trivial fraction of an AUV hotel load budget on long-duration missions.

FPGA-based inference accelerators offer a more efficient alternative: dedicated hardware pipelines for underwater AUV object detection can achieve substantially higher energy efficiency than embedded GPU inference while maintaining real-time frame rates~\cite{qifpga2025}. Model compression techniques such as INT8 quantization and weight pruning further reduce computational overhead on conventional processors. However, these approaches all execute the same multiply-accumulate (MAC) dominated computation graph of conventional ANNs, the per-operation energy of a MAC is bounded by hardware physics and cannot be optimized away in software or by platform choice alone. A qualitatively different computational paradigm is required to reduce inference energy by the amounts that extended AUV missions demand.

\FloatBarrier
\subsection{Spiking Neural Networks for Object Detection}
The sparsity and event-driven nature of SNNs makes them remarkably energy efficient~\cite{rathi2020stdb,spikingdet2021,kimspikingyolo2019}. Kim et al.~\cite{kimspikingyolo2019} were the first to apply SNNs to object detection, converting a tiny-YOLO network via ANN-to-SNN conversion and reporting an estimated 280$\times$ energy reduction on TrueNorth neuromorphic hardware versus GPU-based tiny-YOLO, though without real hardware power measurements. Chakraborty et al.~\cite{spikingdet2021} proposed the Fully Spiking Hybrid Neural Network (FSHNN), combining a RetinaNet backbone with STDP-trained and SGD-trained spiking blocks, fine-tuned via spike timing dependent backpropagation (STDB)~\cite{rathi2020stdb}, achieving 0.426 mAP on COCO. Of particular note, the FSHNN demonstrated improved robustness under small training sets and noisy inputs relative to non-spiking RetinaNet, attributes which are directly relevant to FLS sonar, where labeled data is scarce and speckle noise is pervasive.

More recent work has shifted toward direct training via surrogate gradient descent~\cite{neftci2019surrogate}, where a smooth function approximates the gradient through each binary spike event, yielding a differentiable activation that can be trained via backpropagation, avoiding the accuracy losses of ANN-to-SNN conversion. EMS-YOLO~\cite{su2023emsyolo} demonstrated that a deep directly-trained SNN can achieve competitive COCO performance at $T{=}4$ timesteps. SpikeYOLO~\cite{spikeyolo2024} further advanced the state of the art with integer-valued training and spike-driven inference, reaching 66.2\% mAP@0.5 (48.9\% mAP@0.5:0.95) on COCO. SU-YOLO~\cite{suyolo2025} extended SNN-based detection to underwater optical imagery, but does not address the acoustic domain or energy-constrained AUV deployment. The present work applies SNN-based object detection to FLS sonar imagery for the first time, with explicit energy accounting via the Horowitz model~\cite{horowitz2014}.

\FloatBarrier
\subsection{Neuromorphic Hardware}
IBM's TrueNorth~\cite{merolla2014truenorth} established digital SNN hardware at scale: 4096 cores, one million neurons, and 256 million synapses, achieving 46 billion synaptic operations per second per watt. Intel's Loihi~\cite{davies2018loihi} introduced a manycore neuromorphic processor capable of running SNNs with fully programmable on-chip learning rules. Its successor, Loihi~2~\cite{davies2021loihi2,orchard2021loihi2}, consists of 128 neuron cores per chip with support for custom neuron models via a provided microcode, and scales to one million neurons per chip. Together, these platforms demonstrate that neuromorphic hardware is capable of remarkable energy efficiency and is maturing toward practical deployment, motivating investigation of SNN-based perception pipelines for energy-constrained systems such as AUVs.

Energy estimates throughout this work follow the Horowitz~\cite{horowitz2014} model for 45~nm CMOS: \Emac{} per multiply-accumulate operation and \Eac{} per accumulate-only operation, consistent with the energy model adopted by Kim et al.~\cite{kimspikingyolo2019}. Because spike activations are binary, SNN synaptic operations reduce to AC operations, replacing the MAC operations of conventional ANNs and yielding a theoretical 5.1$\times$ per-operation energy advantage that compounds with the sparse firing rates of biologically-inspired neurons.

\FloatBarrier
\section{Method}
\label{sec:method}

\FloatBarrier
\subsection{Datasets}
\begin{figure}[t]
\centering
\includegraphics[width=\columnwidth]{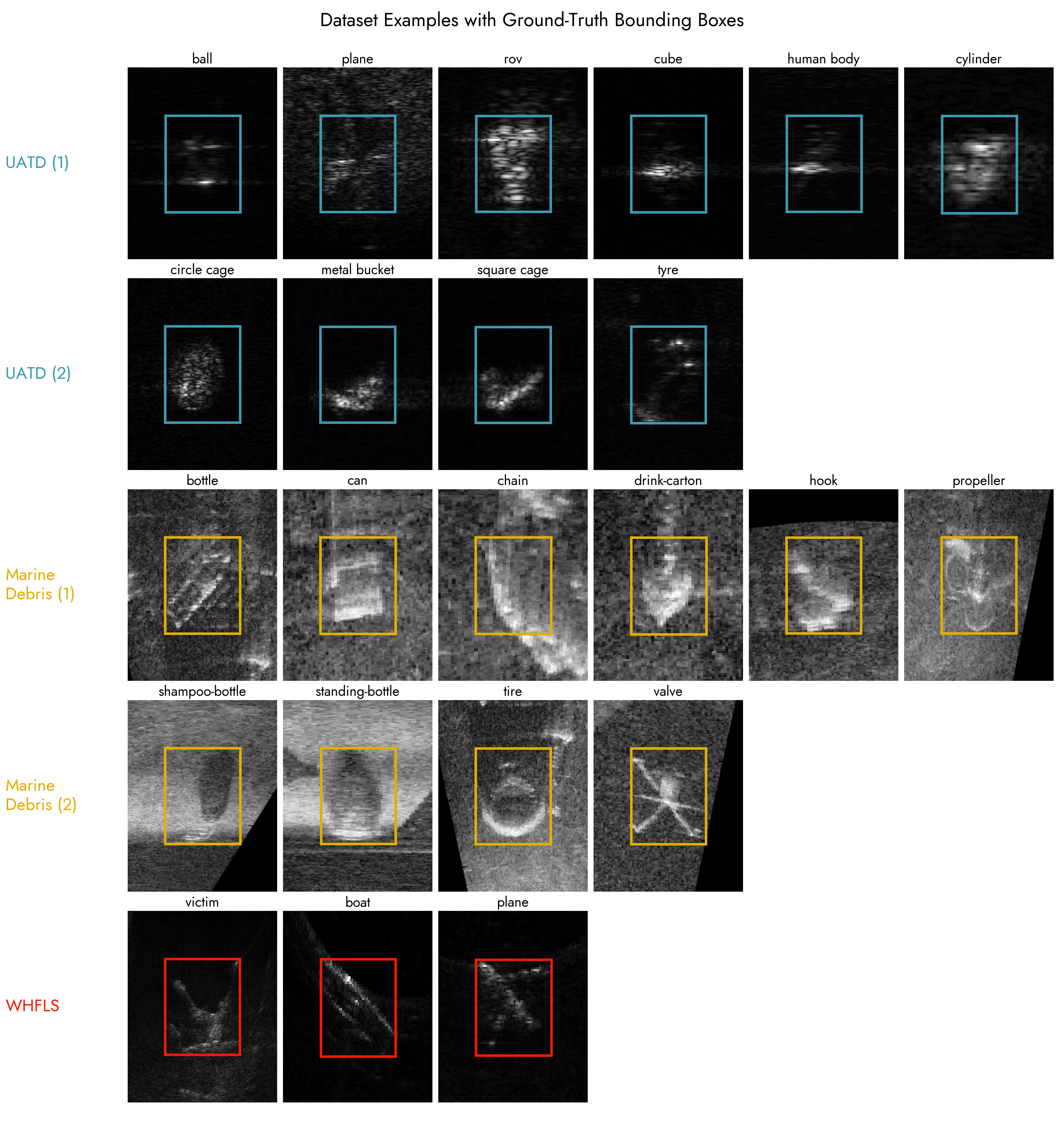}
\caption{Representative FLS images from the three evaluation datasets. Top row: UATD
(10 shape classes, fine-grained discrimination). Middle row: Marine-Debris-FLS (10 classes,
sparse, high-contrast watertank returns). Bottom row: WHFLS (3 classes, victim/boat/plane).
Ground-truth bounding boxes shown in color per dataset.}
\label{fig:dataset_examples}
\end{figure}

Labeled object detection datasets for FLS sonar are scarce. This work evaluates on three publicly available datasets spanning different sonar platforms, environments, and object categories (Fig.~\ref{fig:dataset_examples}).

\textbf{UATD}~\cite{uatd2022} consists of 9200 FLS images collected using a Tritech Gemini 1200ik multibeam FLS in lake and shallow-water environments at ranges of 5--25~m, across two operating frequencies (2900 images at 720~kHz; 6300 at 1200~kHz). Ten shape classes are labeled: ball, plane, cylinder, cube, tyre, ROV, square cage, human body, metal bucket, and circle cage. The official distribution provides separate training and test archives, but the test archives contained empty subdirectories; we therefore apply a random 70/15/15 train/validation/test split, yielding 6440 train / 1380 validation / 1380 test images.

\textbf{Marine-Debris-FLS}~\cite{valdenegro2025marine} was collected using an ARIS Explorer 3000 FLS across three scenarios: a controlled watertank, an underwater turntable, and a flooded quarry. We use only the watertank object detection subset, comprising 1868 full-size FLS images of man-made marine debris across 10 classes: bottle, can, chain, drink carton, hook, propeller, shampoo bottle, standing bottle, tire, and valve. No official split is defined for the object detection task; we use a random 70/15/15 train/validation/test partition (1307/280/281).

\textbf{WHFLS}~\cite{wangwhfls2025} consists of 3752 FLS images collected in open-water conditions using a BlueView M900 sonar at $1024 \times 646$ pixel resolution. Three operationally relevant classes are labeled: victim (submerged diver), boat, and plane. Unlike the controlled laboratory settings of UATD and Marine-Debris-FLS, WHFLS images contain real ocean clutter and environmental noise, making it the most challenging of the three datasets. We follow the author split of 2625 train / 377 validation / 750 test.

\FloatBarrier
\subsection{Network Architectures}
\begin{figure*}[t]
\centering
\includegraphics[width=\textwidth]{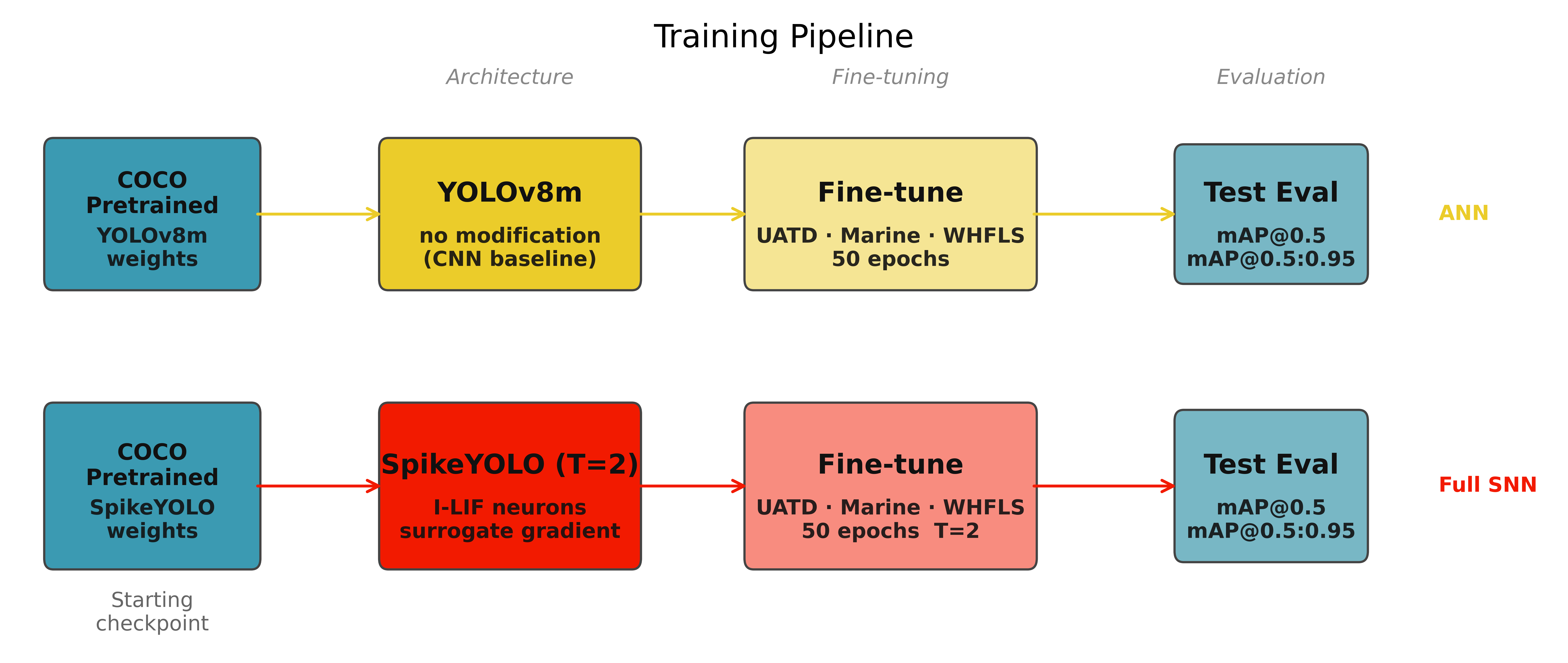}
\caption{SpikeYOLO~\cite{spikeyolo2024} and YOLOv8m~\cite{yolov8} training and evaluation pipeline.
Both models are initialized from a COCO checkpoint and fine-tuned on each FLS dataset.}
\label{fig:pipeline}
\end{figure*}

We evaluate a fully spiking network trained end-to-end with surrogate gradients (Fig.~\ref{fig:pipeline}).

\FloatBarrier
\subsubsection{SpikeYOLO}
SpikeYOLO~\cite{spikeyolo2024} is a fully spiking object detector trained end-to-end via surrogate gradient descent~\cite{neftci2019surrogate}. The architecture follows the YOLOv8 Backbone/Neck/Head structure, replacing the C2f modules with two custom SNN blocks: SNN-Block-1, used in early stages, applies inverted separable convolutions, and SNN-Block-2, used in later stages, applies re-parameterization convolutions for reduced parameter count. Both blocks are adapted from the meta-SNN block of Meta-SpikeFormer~\cite{metaspikeformer} but are fully convolutional with no self-attention. SpikeYOLO uses Integer LIF (I-LIF)~\cite{spikeyolo2024} neurons, which produce integer-valued activations during training to reduce quantization error, but are equivalent to soft-reset LIF neurons running $T \times D$ timesteps at inference, retaining fully accumulate-only operation and compatibility with standard LIF-based neuromorphic frameworks. The model contains 23.1M parameters, implemented via SpikingJelly~\cite{spikingjelly}. We initialize from the publicly released COCO checkpoint (66.2\% mAP@0.5~\cite{spikeyolo2024}) and fine-tune for 50 epochs per dataset at $T{=}2$, retraining from the $T{=}4$ COCO checkpoint to maximize energy efficiency.

\FloatBarrier
\subsubsection{Energy Model}
Inference energy is estimated following the Horowitz~\cite{horowitz2014} model for 45~nm CMOS, as adopted by Kim et al.~\cite{kimspikingyolo2019}. For a conventional ANN, every weight-activation product is a multiply-accumulate:
\begin{equation}
  E_\text{ANN} = \text{MACs} \times E_\text{MAC} \label{eq:ann_energy}
\end{equation}
where $E_\text{MAC} = E_\text{mult} + E_\text{AC} = 3.7 + 0.9 = 4.6$~pJ at 32-bit FP (Table~\ref{tab:horowitz}). YOLOv8m has 70.02~GMACs, giving $E_\text{ANN} = 322$~mJ regardless of dataset.

SpikeYOLO is a fully spiking network in which every synaptic operation is accumulate-only. Its Integer LIF (I-LIF)~\cite{spikeyolo2024} neurons produce integer-valued activations $s \in \{0,1,2,3,4\}$, so the mean activation $\bar{s}$ is an average spike magnitude rather than a binary firing fraction. Total inference energy is:
\begin{equation}
  E_\text{SpikeYOLO} = T \cdot \bar{s} \cdot N_\text{OPS} \cdot E_\text{AC} \label{eq:spikeyolo_energy}
\end{equation}
where $T$ is the number of timesteps and $N_\text{OPS} = 60.93$ is the total synaptic operations per timestep (computed via \texttt{thop} at $T{=}1$).

Because spike rates depend on the input distribution, $\bar{s}$ is measured on each test set rather than taken from COCO literature values (see Table~\ref{tab:spike_rates} in Section~\ref{sec:results_energy}).

\begin{table}[h]
\centering
\caption{Operation energy costs in 45~nm CMOS at 0.9~V~\cite{horowitz2014}.}
\label{tab:horowitz}
\begin{tabular}{lrrr}
\toprule
Operation & 8-bit & 16-bit & 32-bit \\
\midrule
Integer add      & 0.03~pJ & ---     & 0.1~pJ \\
Integer multiply & 0.2~pJ  & ---     & 3.1~pJ \\
FP add           & ---     & 0.4~pJ  & 0.9~pJ \\
FP multiply      & ---     & 1.1~pJ  & 3.7~pJ \\
\bottomrule
\end{tabular}
\end{table}

\FloatBarrier
\subsection{CNN Baselines}
We evaluate several conventional ANN detectors as baselines for comparison against the proposed SNN architectures. YOLOv8m~\cite{yolov8}, with 25.9M parameters, is the primary CNN baseline and shares its architectural basis with the sonar-adapted detector proposed by Zheng et al.~\cite{zheng2024yolov8sonar}. YOLOv8m is initialized from a COCO pretrained checkpoint and fine-tuned for 50 epochs using Ultralytics default hyperparameters. RT-DETR-L~\cite{rtdetr} is a transformer-based detector and the architectural basis of US-DETR~\cite{usdert2025}, and was initialized from a COCO pretrained checkpoint and fine-tuned for 50 epochs.

Faster R-CNN~\cite{fasterrcnn} was benchmarked on UATD by the dataset authors~\cite{uatd2022} using three backbone variants (ResNet-18/50/101). The best result (ResNet-18, 0.839 mAP@0.5) is included here as a representative two-stage detection baseline. On Marine-Debris-FLS, we additionally report published results for YOLO-SONAR~\cite{wangwhfls2025}, a YOLOv7-based network with attention modules tailored to FLS imagery, and for the Fast R-CNN architecture of Karimanzira et al.~\cite{karimanzira2020}, neither was retrained here.

\FloatBarrier
\subsection{Training Protocol}

All spiking and CNN models were fine-tuned per dataset from COCO-pretrained checkpoints on a single NVIDIA L40S GPU using Ultralytics~8.4.x and SpikingJelly~\cite{spikingjelly}. Table~\ref{tab:training} summarises the per-model configuration. Faster~R-CNN~\cite{uatd2022}, YOLO-SONAR~\cite{wangwhfls2025}, and the Fast R-CNN of Karimanzira et al.~\cite{karimanzira2020} results are taken from the cited works and were not retrained here.

SpikeYOLO is initialized from the publicly released $T{=}4$ COCO checkpoint and fine-tuned directly at $T{=}2$.

\begin{table}[h]
\centering
\caption{Training configuration for all retrained models.}
\label{tab:training}
\begin{tabular}{lllr}
\toprule
Model & Init.\ checkpoint & Optimizer / LR / schedule & Epochs \\
\midrule
YOLOv8m~\cite{yolov8}          & COCO           & SGD, lr\,=\,0.01, cosine                & 50 \\
RT-DETR-L~\cite{rtdetr}        & COCO           & Ultralytics defaults                      & 50 \\
SpikeYOLO~\cite{spikeyolo2024} & COCO ($T{=}4$) & AdamW, lr\,=\,$10^{-4}$, cosine          & 50 \\
\bottomrule
\end{tabular}
\smallskip\\
{\footnotesize SpikeYOLO fine-tuned at $T{=}2$ from $T{=}4$ COCO checkpoint.}
\end{table}

\FloatBarrier
\section{Results}
\label{sec:results}

\FloatBarrier
\subsection{UATD Detection Performance}
\label{sec:results_uatd}

\begin{table}[t]
\centering
\caption{UATD detection results. \textbf{Bold}: best per column. Dashes: values not published in literature for architecture.
All mAP@0.5:0.95 values are means over 3 independent seeds.}
\label{tab:uatd}
\footnotesize
\begin{tabular}{llrrr}
\toprule
Model & Type & \mAPf & \mAP & Energy\,(mJ) \\
\midrule
Faster R-CNN~\cite{uatd2022}     & CNN         & 0.839 & —                       & — \\
RT-DETR-L~\cite{rtdetr}          & Transformer & \textbf{0.979} & {\boldmath$0.642\pm0.005$} & 476 \\
YOLOv8m~\cite{yolov8}            & CNN         & 0.969 & $0.575\pm0.004$          & 322 \\
\midrule
SpikeYOLO (T=2)~\cite{spikeyolo2024} & Fully spiking & 0.964 & $0.529\pm0.005$ & \textbf{97} \\
\bottomrule
\end{tabular}
\end{table}

SpikeYOLO trades a modest accuracy reduction for substantial energy savings on UATD (Table~\ref{tab:uatd}). SpikeYOLO (T=2) achieves 0.529~$\pm$~0.005 mAP@0.5:0.95, 0.046 below YOLOv8m's 0.575~$\pm$~0.004, at 3.3$\times$ lower theoretical compute energy (97 vs 322~mJ). SpikeYOLO outperforms the Faster R-CNN baseline~\cite{uatd2022} in mAP@0.5. RT-DETR-L achieves the highest accuracy but at 1.48$\times$ higher energy than YOLOv8m (476 vs 322~mJ) and 4.9$\times$ higher than SpikeYOLO (476 vs 97~mJ). Low variance across three seeds confirms the results are not seed-dependent.

\begin{figure}[t]
\centering
\includegraphics[width=\columnwidth]{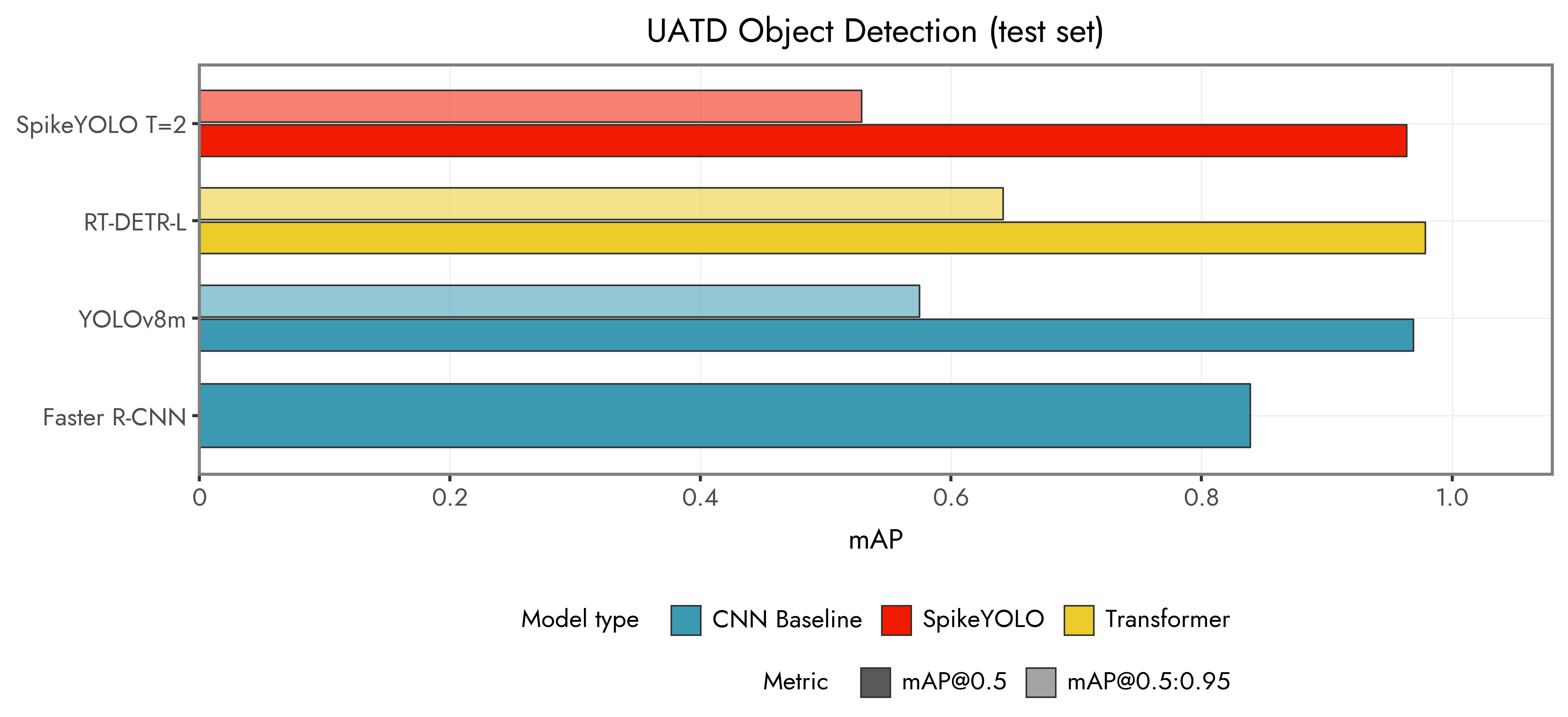}
\caption{UATD detection accuracy (mAP@0.5 and mAP@0.5:0.95) by network. SNN achieves competitive accuracy.}
\label{fig:uatd_comparison}
\end{figure}

\FloatBarrier
\subsection{Marine Debris Detection}
\label{sec:results_marine}

\begin{table}[t]
\centering
\caption{Marine-debris-fls detection results.
\textbf{Bold}: best per column. Dashes: values not published in literature for architecture.
All mAP@0.5:0.95 values are means over 3 independent seeds.}
\label{tab:marine}
\footnotesize
\begin{tabular}{llrrr}
\toprule
Model & Type & \mAPf & \mAP & Energy\,(mJ) \\
\midrule
YOLO-SONAR~\cite{wangwhfls2025}  & CNN         & 0.953 & 0.667 & — \\
Fast R-CNN~\cite{karimanzira2020} & CNN         & 0.967 & 0.759 & — \\
YOLOv8m~\cite{yolov8}             & CNN         & 0.987 & $0.801\pm0.006$ & 322 \\
RT-DETR-L~\cite{rtdetr}           & Transformer & \textbf{0.989} & {\boldmath$0.819\pm0.008$} & 476 \\
\midrule
SpikeYOLO (T=2)~\cite{spikeyolo2024} & Fully spiking & 0.987 & $0.781\pm0.008$ & \textbf{73} \\
\bottomrule
\end{tabular}
\end{table}

SpikeYOLO matches YOLOv8m on mAP@0.5 (0.987) and approaches RT-DETR-L (0.989) while using 4.4$\times$ less energy (73 vs 322~mJ) (Table~\ref{tab:marine}). On mAP@0.5:0.95, SpikeYOLO ($0.781\pm0.008$) falls below YOLOv8m ($0.801\pm0.006$) by 0.020 and below RT-DETR-L ($0.819\pm0.008$) by 0.038, suggesting the gap is in box localization precision rather than detection. SpikeYOLO outperforms the YOLO-SONAR and Fast R-CNN baselines on both metrics. Low variance across three seeds confirms the results are not seed-dependent.

\begin{figure}[t]
\centering
\includegraphics[width=\columnwidth]{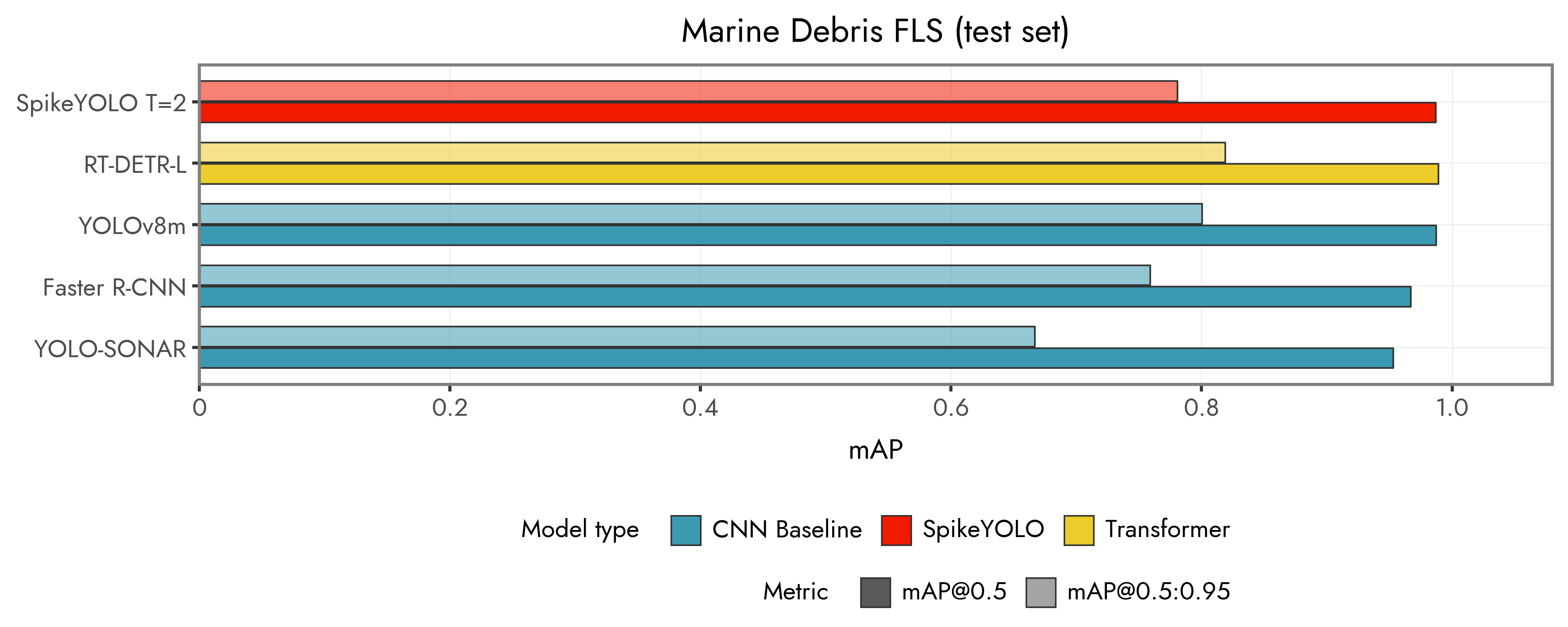}
\caption{Marine-debris-fls detection accuracy (mAP@0.5 and mAP@0.5:0.95) by network. SpikeYOLO matches YOLOv8m on mAP@0.5.}
\label{fig:marine_comparison}
\end{figure}

\FloatBarrier
\subsection{WHFLS Detection Performance}
\label{sec:results_whfls}

\begin{table}[t]
\centering
\caption{WHFLS detection results. \textbf{Bold}: best per column. Dashes: values not published in literature for architecture.
All mAP@0.5:0.95 values are means over 3 independent seeds.}
\label{tab:whfls}
\footnotesize
\begin{tabular}{llrrr}
\toprule
Model & Type & \mAPf & \mAP & Energy\,(mJ) \\
\midrule
YOLOv8m~\cite{yolov8}                   & CNN           & 0.9949 & $0.872\pm0.004$ & 322 \\
RT-DETR-L~\cite{rtdetr}                 & Transformer   & \textbf{0.9950} & {\boldmath$0.893\pm0.001$} & 476 \\
\midrule
SpikeYOLO (T=2)~\cite{spikeyolo2024}    & Fully spiking & \textbf{0.9950} & $0.861\pm0.001$ & \textbf{52} \\
\bottomrule
\end{tabular}
\end{table}

SpikeYOLO and RT-DETR-L both achieve mAP@0.5 of 0.9950 versus YOLOv8m's 0.9949 (Table~\ref{tab:whfls}), with SpikeYOLO using 6.2$\times$ less energy than YOLOv8m (52 vs 322~mJ) and 9.2$\times$ less than RT-DETR-L (52 vs 476~mJ). On mAP@0.5:0.95, SpikeYOLO ($0.861\pm0.001$) falls 0.011 below YOLOv8m ($0.872\pm0.004$) and 0.032 below RT-DETR-L ($0.893\pm0.001$). The near-perfect mAP@0.5 scores reflect WHFLS's three visually distinct classes, while the mAP@0.5:0.95 gap indicates that the SNN is less precise at localizing objects. Low variance across three seeds confirms the results are not seed-dependent.

\begin{figure}[t]
\centering
\includegraphics[width=\columnwidth]{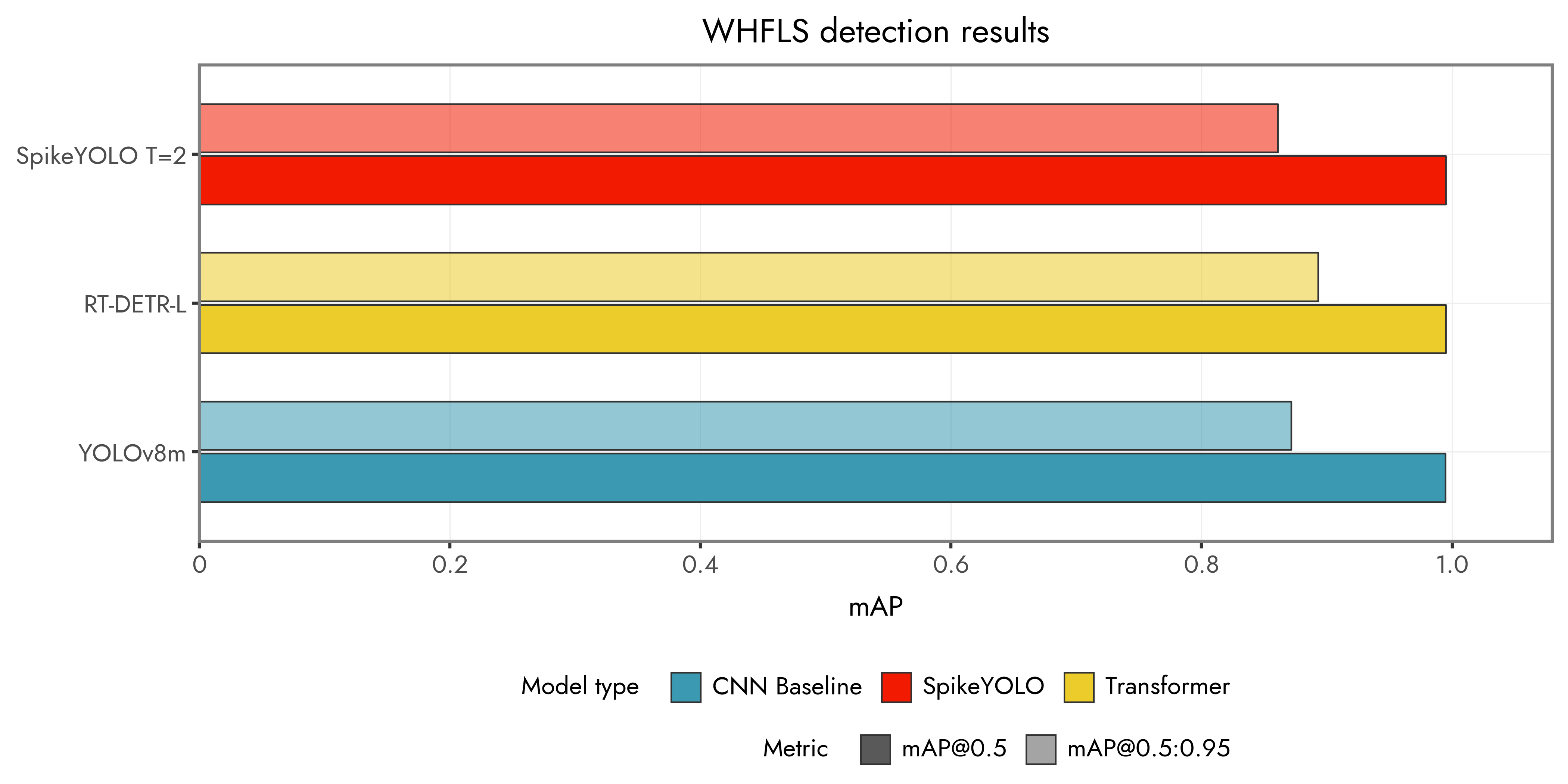}
\caption{WHFLS detection accuracy (mAP@0.5 and mAP@0.5:0.95) by network. SpikeYOLO beats YOLOv8m on mAP@0.5 (0.9950).}
\label{fig:whfls_comparison}
\end{figure}

\FloatBarrier
\subsection{Inference Energy}
\label{sec:results_energy}

\begin{figure}[t]
\centering
\includegraphics[width=\columnwidth]{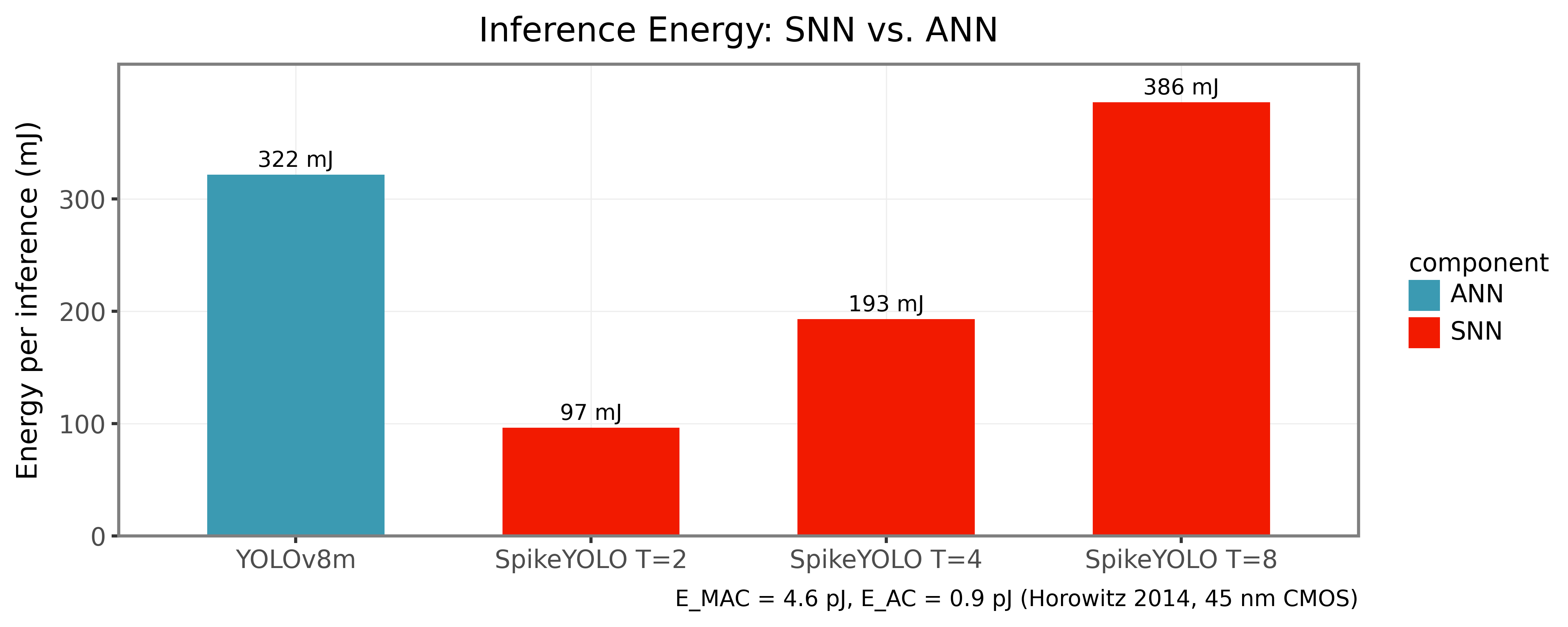}
\caption{Inference energy breakdown per model.
SpikeYOLO uses accumulate-only operations,
achieving 3.3$\times$ lower theoretical energy than YOLOv8m at $T{=}2$.}
\label{fig:energy}
\end{figure}

\begin{table}[t]
\centering
\caption{Model efficiency comparison on UATD.
Energy: theoretical CMOS estimate (\Emac, \Eac, \cite{horowitz2014}).}
\label{tab:efficiency}
\begin{tabular}{lrrr}
\toprule
Model & Params\,(M) & Energy\,(mJ) & \mAP \\
\midrule
YOLOv8m~\cite{yolov8}                   & 25.9 & 322            & $0.575\pm0.004$ \\
\midrule
SpikeYOLO (T=2)~\cite{spikeyolo2024}    & 23.1 & 97             & $0.529\pm0.005$ \\
\bottomrule
\end{tabular}
\end{table}

\begin{table}[t]
\centering
\caption{Mean spike activation $\bar{s}$ (mean I-LIF integer output per neuron per
  timestep, range 0--4) measured on each FLS test set at $T{=}2$.}
\label{tab:spike_rates}
\begin{tabular}{llr}
\toprule
Model & Dataset & $\bar{s}$ \\
\midrule
\multirow{3}{*}{SpikeYOLO T=2}     & UATD          & 0.880 \\
                                    & Marine Debris & 0.668 \\
                                    & WHFLS         & 0.476 \\
\bottomrule
\end{tabular}
\end{table}

We compute energy estimates using the model outlined by Horowitz et al.~\cite{horowitz2014} and commonly used to calculate energy usage in neuromorphic literature~\cite{kimspikingyolo2019,spikeyolo2024}. For YOLOv8m we calculate the energy usage of the network according to Eq.~\eqref{eq:ann_energy}, with MACs measured via \texttt{thop} at 69.9~GMACs and $E_\text{MAC}=4.6$~pJ, yielding 322~mJ per forward pass. For SpikeYOLO, the fully spiking network, we apply Eq.~\eqref{eq:spikeyolo_energy} with $N_\text{OPS}=60.93$ and $\bar{s}$ varying by dataset (Table~\ref{tab:spike_rates}). A summary of model efficiency on UATD is provided in Table~\ref{tab:efficiency}.

\FloatBarrier
\subsection{Timestep Ablation}
\label{sec:t_ablation}

\begin{figure}[t]
\centering
\includegraphics[width=\columnwidth]{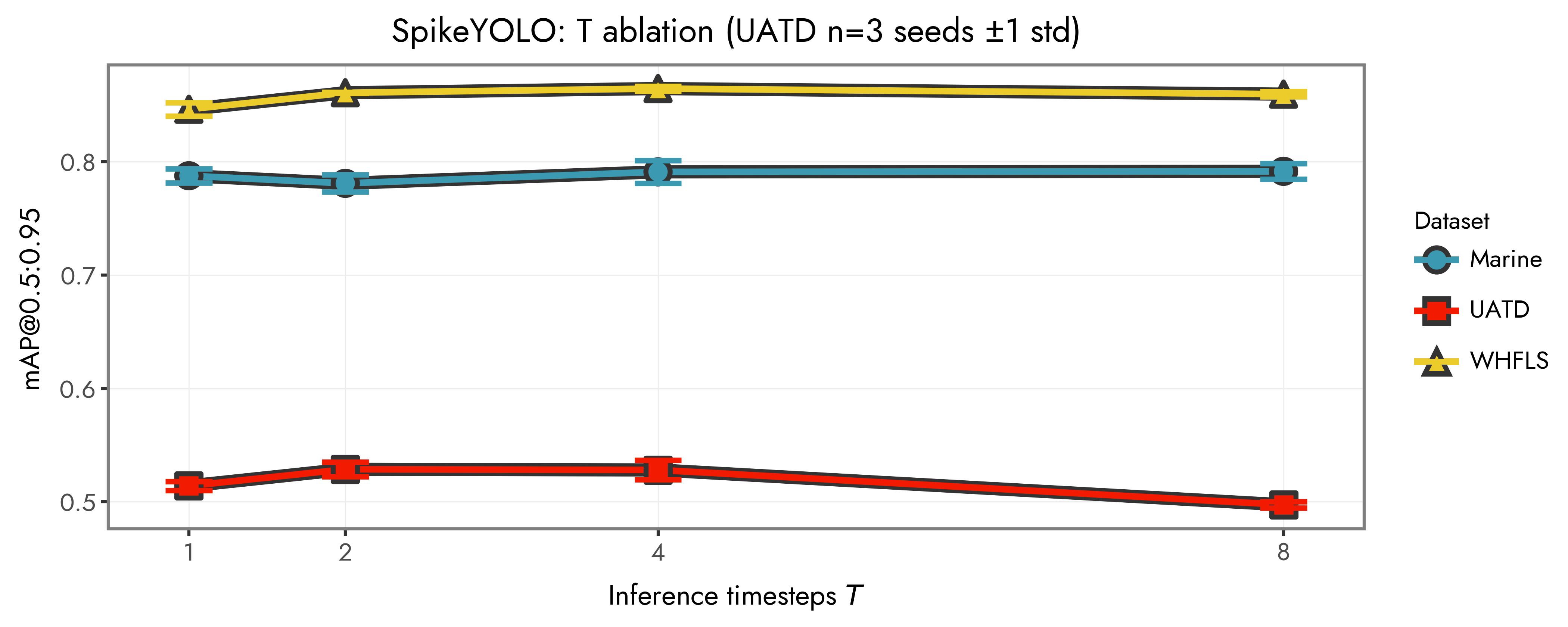}
\caption{mAP@0.5:0.95 vs.\ inference timesteps $T$ for SpikeYOLO
on all three FLS datasets. Accuracy is relatively flat across
$T \in \{2,4,8\}$; energy scales linearly with $T$, $T{=}2$ is Pareto-optimal.}
\label{fig:t_ablation}
\end{figure}

Inference energy scales linearly with $T$ (Eq.~\eqref{eq:spikeyolo_energy}). We evaluate SpikeYOLO~\cite{spikeyolo2024} at $T \in \{1, 2, 4, 8\}$ on all three datasets by retraining from the COCO checkpoint at each $T$. As shown in Fig.~\ref{fig:t_ablation}, accuracy at $T{=}1$ drops noticeably relative to $T \geq 2$, while $T \in \{2, 4, 8\}$ remain flat across all datasets. The measured spike rates (Table~\ref{tab:spike_rates}) confirm that even at $T{=}2$ the model accumulates sufficient rate-coded signal from FLS imagery's sparse structure. Since $T{=}2$ achieves the same accuracy as larger $T$ at half the energy of $T{=}4$, we use $T{=}2$ for all accuracy and energy comparisons in this paper.

\FloatBarrier
\subsection{Robustness to Speckle Noise}
\label{sec:results_noise}

\begin{figure}[t]
\centering
\includegraphics[width=\columnwidth]{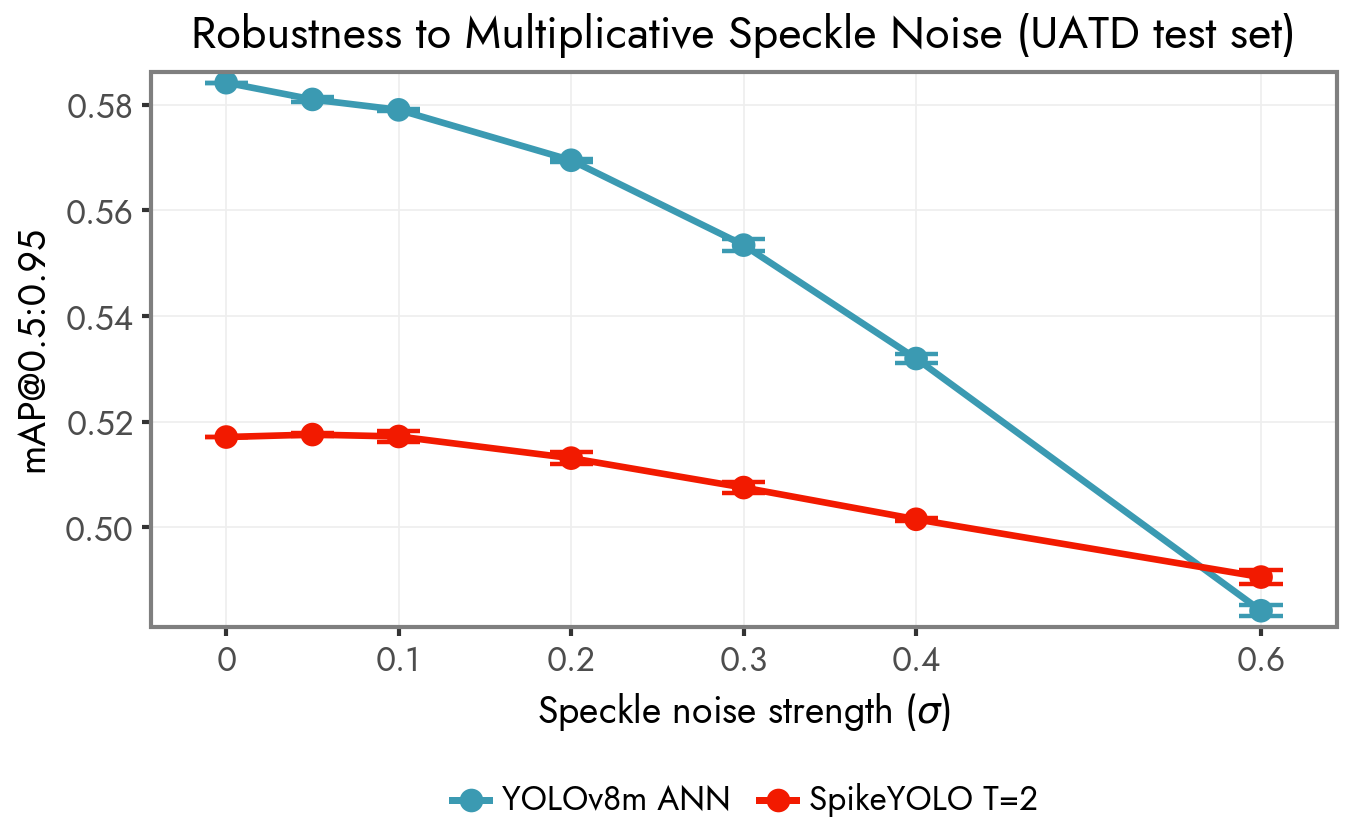}
\caption{mAP@0.5:0.95 under multiplicative speckle noise,
evaluated on UATD test set without retraining.
SpikeYOLO~\cite{spikeyolo2024} degrades 3.0\% at $\sigma{=}0.4$ vs.\ 8.9\% for
YOLOv8m~\cite{yolov8}. At $\sigma{=}0.6$ the fully-spiking model outperforms the ANN
baseline outright.}
\label{fig:noise}
\end{figure}

Multiplicative speckle noise is the primary source of noise in FLS imagery. To
assess the noise robustness of the networks we apply multiplicative speckle
noise in accordance to Eq.~\eqref{eq:speckle} to the images in the UATD test
set and evaluate without retraining. From $\sigma=0$ to $\sigma=0.4$ YOLOv8m's
mAP@0.5:0.95 falls 8.9\% while SpikeYOLO falls only 3.0\% (Fig.~\ref{fig:noise}). At $\sigma=0.6$, SpikeYOLO outperforms YOLOv8m outright (0.491 vs.\ 0.484~\mAP{}). SpikeYOLO performs
significantly better on noisy data, consistent with the noise-filtering
property of LIF neurons noted by \cite{spikingdet2021}.

\begin{equation}
  \label{eq:speckle}
  I_\text{noisy} = I \cdot (1 + \sigma \mathcal{N}(0,1)),
   \sigma \in \{0, 0.05, 0.1, 0.2, 0.3, 0.4, 0.6\}
\end{equation}

\FloatBarrier
\subsection{Low-Data Regime}
\label{sec:results_lowdata}

\begin{table}[t]
\centering
\caption{Low-data ablation: mAP@0.5:0.95 on UATD test set at reduced
training fractions. Mean\,$\pm$\,std over 2 random seeds. 100\% results from 3-seed
retrains (Section~\ref{sec:results_uatd}).}
\label{tab:lowdata}
\begin{tabular}{lrr}
\toprule
Training fraction & YOLOv8m~\cite{yolov8} & SpikeYOLO T=2~\cite{spikeyolo2024} \\
\midrule
25\%  & $0.449 \pm 0.011$ & $\mathbf{0.475} \pm 0.006$ \\
50\%  & $0.491 \pm 0.009$ & $\mathbf{0.513} \pm 0.005$ \\
100\% & $0.575 \pm 0.004$ & $0.529 \pm 0.005$           \\
\bottomrule
\end{tabular}
\smallskip\\
{\footnotesize All values are mean mAP@0.5:0.95 over 3 seeds (100\%) or 2 seeds (25\%, 50\%).}
\end{table}

\begin{figure}[t]
\centering
\includegraphics[width=\columnwidth]{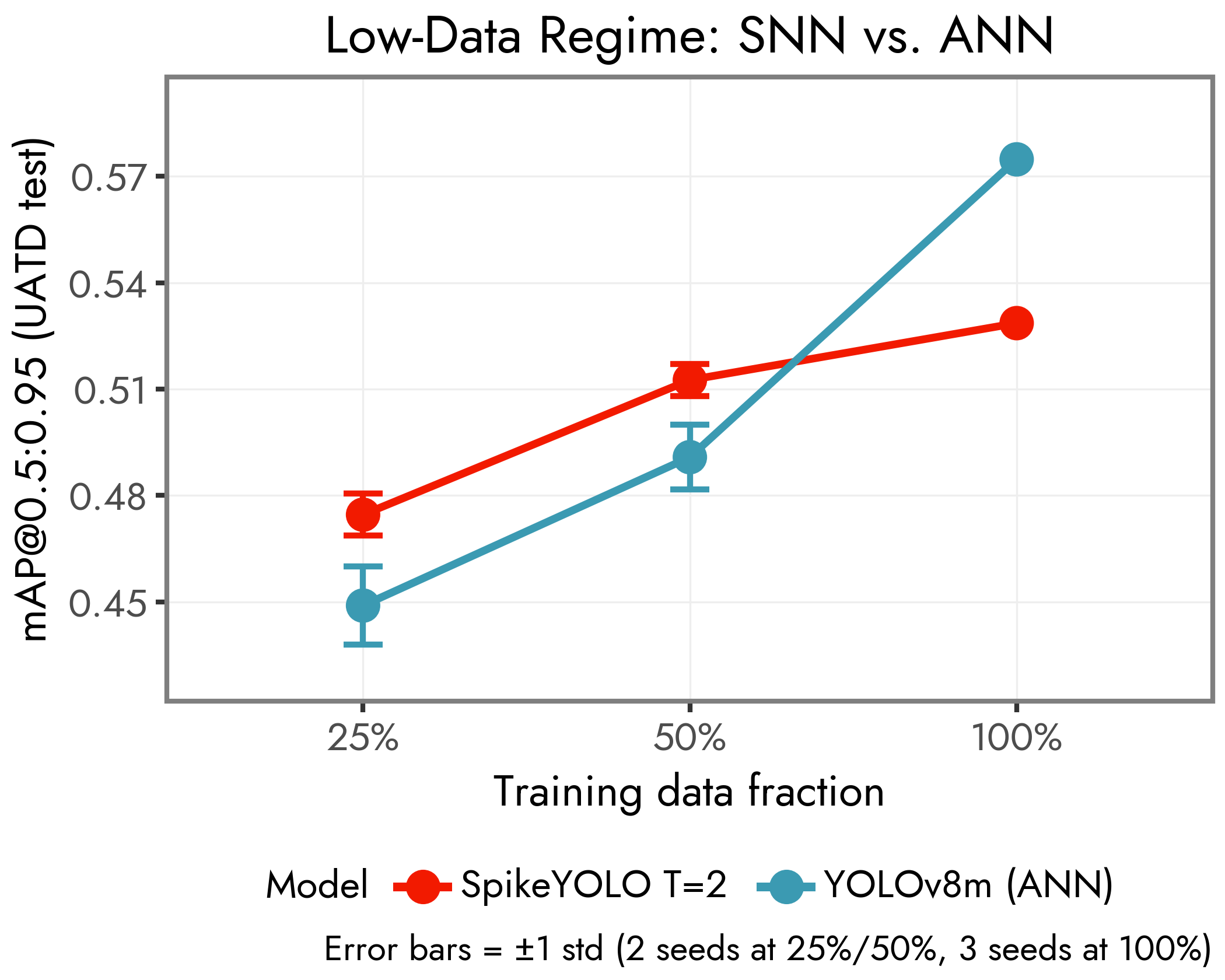}
\caption{mAP@0.5:0.95 vs.\ training data fraction for YOLOv8m~\cite{yolov8} and
SpikeYOLO~\cite{spikeyolo2024} T=2 on UATD~\cite{uatd2022}. SpikeYOLO leads at
25\% and 50\% data; YOLOv8m recovers at 100\%.}
\label{fig:lowdata}
\end{figure}

Quality labeled FLS imagery is often expensive and difficult to obtain, so training data is frequently sparse. In order to assess the robustness of each network architecture to limited training data we randomly subsampled the UATD training images and retrained each network according to the procedure outlined in Section~\ref{sec:method}. For both 25\% and 50\% of the data SpikeYOLO~\cite{spikeyolo2024} outperforms YOLOv8m (5.8\% greater mAP@0.5:0.95 at 25\% and 4.4\% greater mAP@0.5:0.95 at 50\%, Table~\ref{tab:lowdata}, Fig.~\ref{fig:lowdata}), following the established pattern in the literature that fully spiking neural networks often perform better than their ANN counterparts when training availability is limited.

\FloatBarrier
\subsection{Per-Class and Qualitative Analysis}
\label{sec:results_perclass}

\begin{figure}[t]
\centering
\includegraphics[width=\columnwidth]{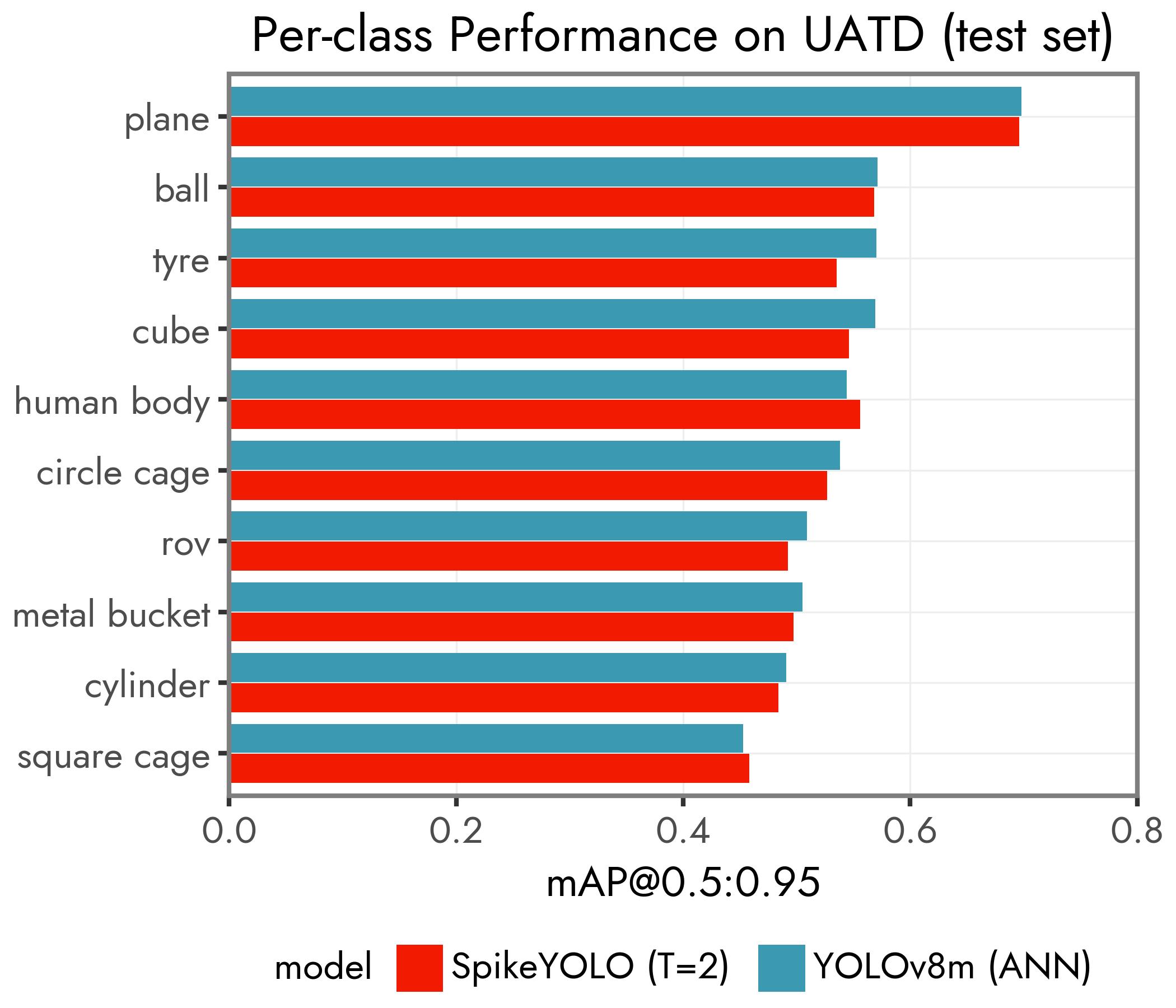}
\caption{Per-class mAP@0.5:0.95 on UATD~\cite{uatd2022}. Plane and ball achieve
the highest accuracy across all models; square cage and cylinder are consistently
challenging due to complex geometry.}
\label{fig:perclass}
\end{figure}

\begin{figure*}[t]
\centering
\includegraphics[width=\textwidth,height=0.85\textheight,keepaspectratio]{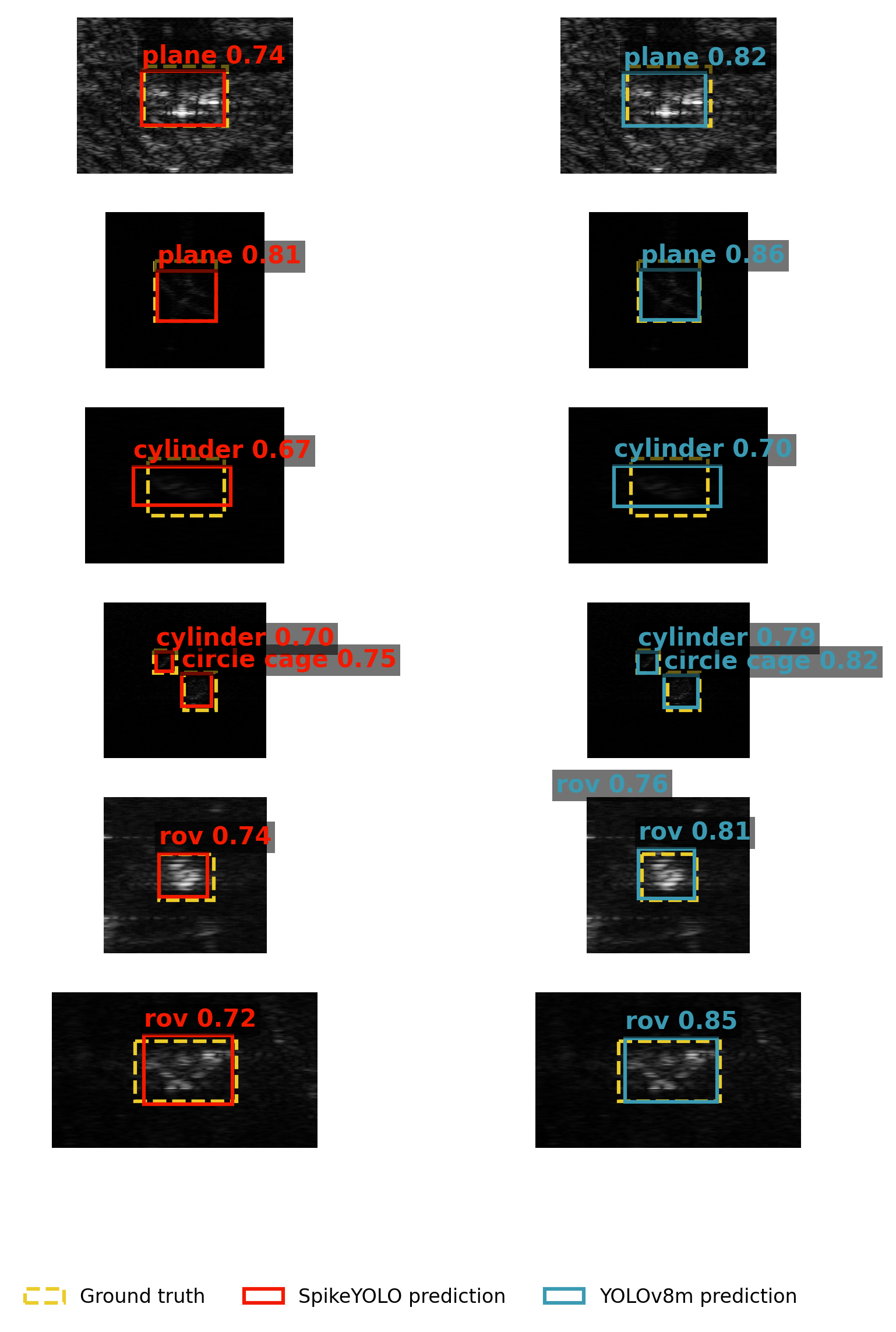}
\caption{Qualitative detection results on UATD~\cite{uatd2022} test images.
Left: SpikeYOLO~\cite{spikeyolo2024} T=2. Right: YOLOv8m~\cite{yolov8}.
Dashed white: ground truth. Solid: predictions with class label and confidence score.
Rows: plane, cylinder, circle cage, ROV.}
\label{fig:qualitative}
\end{figure*}

Fig.~\ref{fig:perclass} shows a consistent cross-model pattern of per-class accuracy differences on UATD. Across all models, plane achieves the highest AP@0.5:0.95, followed by ball and tyre. Square cage and cylinder are the lowest, with ROV close behind, and the remaining classes in between. Because this ranking is broadly consistent regardless of architecture, the difficulty appears to be driven by inherent object properties in sonar imaging rather than model capability. WHFLS shows no significant per-class differences in accuracy.

Fig.~\ref{fig:qualitative} shows representative detection examples from UATD. Both models localize the plane reliably. Performance degrades on ROV, cylinder, and square cage, where both models produce lower-confidence detections and occasional missed objects.

\FloatBarrier
\section{Discussion}
\label{sec:discussion}

Marine Debris FLS and WHFLS exhibit sparse, high-contrast returns, and on these datasets SNNs achieve comparable or greater accuracy than the CNN baselines and occasionally match the transformer-based RT-DETR-L while using less energy. UATD, however, comprises a greater number of classes with more complex acoustic signatures, and SNNs score close but below YOLOv8m while still requiring less energy. Spiking neural networks appear to perform best when the objects to be detected have simple, high-contrast returns, but remain competitive even on more complex tasks.
SpikingYOLOv8, a hybrid ANN--SNN architecture using STDB threshold-balancing conversion~\cite{rathi2020stdb}, was also evaluated but achieved only $0.359\pm0.037$~\mAP{} on UATD after fine-tuning (mean over 3 seeds), substantially below both SpikeYOLO (0.529) and YOLOv8m (0.575). This suggests that ANN-to-SNN conversion is less effective for FLS detection than end-to-end surrogate-gradient training.
Kim et al.'s~\cite{kimspikingyolo2019} estimates and those we make above assume that weights are stored in an energy-efficient on-chip memory such as SRAM and that memory-movement costs are not relevant. However, Horowitz~\cite{horowitz2014} reports that off-chip memory costs can be significant, with a 64-bit DRAM access costing 1.3--2.6~nJ, or $\sim$20--41~pJ/bit, as opposed to 100~pJ for a 64-bit access to a 1~MB SRAM cache, or $\sim$1.6~pJ/bit. SpikeYOLO has 23.1M parameters that occupy $\sim$92~MB and YOLOv8m has 25.9M parameters that occupy $\sim$104~MB, more than we can necessarily expect to be able to store in on-chip memory. If for simplicity we assume a likely overly optimistic scenario of a single read of all parameters from DRAM for each inference, this costs 15--30~mJ for SpikeYOLO and 17--34~mJ for YOLOv8m, increasing the energy cost of an inference on SpikeYOLO to 112--127~mJ and on YOLOv8m to 339--356~mJ, reducing the 3.3$\times$ energy advantage on UATD to a smaller but still significant 2.8--3.0$\times$ energy advantage. Because this DRAM overhead is fixed by parameter count rather than by dataset, it narrows SpikeYOLO's advantage on Marine-Debris-FLS and WHFLS as well (Table~\ref{tab:dram_datasets}). However, quantization of network parameters along with growing on-chip memory caches opens the possibility of fitting the described networks on chip. Deployment on modern neuromorphic hardware such as Intel's Loihi 2 is likely to yield even lower energy consumption, true energy measurements are left to future work pending hardware access. Evaluation is limited to three FLS datasets, with broader benchmarking constrained by the scarcity of publicly available labeled FLS data, with many datasets being proprietary to the defense and energy sectors.

\begin{table}[t]
\centering
\caption{SpikeYOLO vs.\ YOLOv8m energy, compute-only vs.\ including the estimated single-read DRAM weight-fetch overhead (15--34~mJ, fixed by parameter count rather than dataset).}
\label{tab:dram_datasets}
{\footnotesize
\setlength{\tabcolsep}{3pt}
\begin{tabular}{lrrrrrr}
\toprule
 & \multicolumn{2}{c}{SpikeYOLO (mJ)} & \multicolumn{2}{c}{YOLOv8m (mJ)} & \multicolumn{2}{c}{Ratio} \\
\cmidrule(lr){2-3} \cmidrule(lr){4-5} \cmidrule(lr){6-7}
Dataset & compute & +DRAM & compute & +DRAM & compute-only & +DRAM \\
\midrule
UATD              & 97 & 112--127 & 322 & 339--356 & 3.3$\times$ & 2.8--3.0$\times$ \\
Marine-Debris-FLS & 73 & 88--103  & 322 & 339--356 & 4.4$\times$ & 3.5--3.9$\times$ \\
WHFLS             & 52 & 67--82   & 322 & 339--356 & 6.2$\times$ & 4.3--5.1$\times$ \\
\bottomrule
\end{tabular}}
\end{table}

\FloatBarrier
\section{Conclusion}
\label{sec:conclusion}

AUVs operate off fixed batteries and are therefore energy constrained. Hotel loads, including sonar processing, can often consume hundreds of watts of power. Spiking neural networks paired with neuromorphic processing chips provide a low-energy solution to sonar object detection. SpikeYOLO achieves a mAP@0.5:0.95 of 0.529 on UATD, 0.046 below YOLOv8m's 0.575, at 3.3$\times$ lower theoretical inference energy (97 vs.\ 322~mJ); scores within 0.011 of YOLOv8m on WHFLS (0.861 vs.\ 0.872~mAP@0.5:0.95) at the same energy advantage; and scores within 0.020 of YOLOv8m on Marine-Debris-FLS (0.781 vs.\ 0.801~mAP@0.5:0.95) while using less energy. Sparser, higher-contrast, edge-dominated tasks such as WHFLS and Marine-Debris-FLS appear better suited to SNN deployment than tasks with more complex returns such as UATD, although SNNs still achieve similar performance to CNNs on UATD with lower energy usage. SpikeYOLO also shows less performance degradation with the addition of speckle noise (3.0\% degradation vs.\ 8.9\% for YOLOv8m at $\sigma{=}0.4$, outperforming YOLOv8m outright at $\sigma{=}0.6$), and degrades less than YOLOv8m when trained on random subsamples of the training data.

\section*{Acknowledgments}
Partial GPU compute was provided via the Neuroscience Gateway~\cite{nsg2013} (NSG), including access to the Expanse cluster at the San Diego Supercomputer Center (SDSC).

\bibliographystyle{elsarticle-num}
\bibliography{references}

\end{document}